\documentclass{article}
\usepackage{ijcai24}

\usepackage{times}
\usepackage{soul}
\usepackage{url}
\usepackage[hidelinks]{hyperref}
\usepackage[utf8]{inputenc}
\usepackage[small]{caption}
\usepackage{graphicx}
\usepackage{amsmath}
\usepackage{amsthm}
\usepackage{booktabs}
\usepackage{algorithm}
\usepackage{algorithmic}
\usepackage[switch]{lineno}

\usepackage{amssymb}
\usepackage{amsfonts}       % blackboard math symbols

\usepackage{multirow}
\usepackage{doi}

\usepackage[utf8]{inputenc} % allow utf-8 input
\usepackage[T1]{fontenc}    % use 8-bit T1 fonts
\usepackage{hyperref}       % hyperlinks
\usepackage{booktabs}       % professional-quality tables
\usepackage{nicefrac}       % compact symbols for 1/2, etc.
\usepackage{microtype}      % microtypography

\usepackage[capitalize]{cleveref}
\crefname{section}{Sec.}{Secs.}
\Crefname{section}{Section}{Sections}
\Crefname{table}{Table}{Tables}
\crefname{table}{Tab.}{Tabs.}
\crefname{figure}{Fig.}{figures}
\Crefname{figure}{Fig.}{Figures}
\newcommand{\nom}{StegaFormer}
\newcommand{\opme}{OPME}
\newcommand{\opmefull}{Order-Preserving Message Encoder}
\newcommand{\opmd}{OPMD}
\newcommand{\opmdfull}{Order-Preserving Message Decoder}
\newcommand{\gmif}{GMIF}
\newcommand{\gmiffull}{Global Message-Image Fusion}
\newcommand{\wmsa}{W-MSA}
\newcommand{\wmsafull}{Windowed Multi-Head Self-Attention Layer}
\newcommand{\msa}{MSA}

\newcommand{\ie}{\textit{i}.\textit{e}.}

\usepackage{color}
\usepackage[table]{xcolor}
\definecolor{orange}{rgb}{1.0, 0.5, 0.0}

\title{Effective Message Hiding with Order-Preserving Mechanisms}

\author{
Gao Yu$^1$~ Qiu Xuchong$^{1}$\thanks{the corresponding author {\tt\small xuchong.qiu@cn.bosch.com}}~ Ye Zihan$^2$\\
\affiliations
$^{1}$Bosch Corporate Research\\
$^2$Xi’an Jiaotong-Liverpool University\\
\emails
\{yu.gao2, xuchong.qiu\}@cn.bosch.com,
Zihan.Ye22@student.xjtlu.edu.cn\\
}
\begin{document}

\maketitle

% 1. StegaFormer: Efficient Transformer for Message Hidding

% 2. Rethinking/Exploring MLPs/Transformers for Message Hidding

% 3. Towards Effective Message Hidding via MLPs/Transformers

\begin{abstract}
Message hiding, a technique that conceals secret message bits within a cover image, aims to achieve an optimal balance among message capacity, recovery accuracy, and imperceptibility. While convolutional neural networks have notably improved message capacity and imperceptibility, achieving high recovery accuracy remains challenging. This challenge arises because convolutional operations struggle to preserve the sequential order of message bits and effectively address the discrepancy between these two modalities. To address this, we propose StegaFormer, an innovative MLP-based framework designed to preserve bit order and enable global fusion between modalities. Specifically, StegaFormer incorporates three crucial components: Order-Preserving Message Encoder (OPME), Decoder (OPMD) and Global Message-Image Fusion (GMIF). OPME and OPMD aim to preserve the order of message bits by segmenting the entire sequence into equal-length segments and incorporating sequential information during encoding and decoding. Meanwhile, GMIF employs a cross-modality fusion mechanism to effectively fuse the features from the two uncorrelated modalities.
Experimental results on the COCO and DIV2K datasets demonstrate that StegaFormer surpasses existing state-of-the-art methods in terms of recovery accuracy, message capacity, and imperceptibility. Our code is released in https://github.com/boschresearch/Stegaformer.
\end{abstract}

\section{Introduction}
\label{sec:intro}
%% desicribe the task and why it's important as %%
Image steganography is a technique that conceals confidential information within a publicly accessible image ~\cite{mielikainen2006lsb,pevny2010using,zhang2019steganogan,tan2021channel}.
In this study, we focus on message hiding.
It has a wide range of applications such as copy-right protection, light field messaging~\cite{Wengrowski2019lfm}, virtual reality and augmented reality applications~\cite{2019stegastamp} and face anonymization~\cite{kishore2021fixed}.
The goal of it is embedding secret message bits within a cover image to produce a new image (called stego image). 
Ensuring a balanced trade-off among message capacity, recovery accuracy, and imperceptibility is the task's core.
Generally, message capacity is measured by bits per pixel (BPP).
Recovery accuracy indicates the message fidelity after being processed by these methods. Imperceptibility is minimizing the perceptible differences between the cover image and the stego image.

\begin{figure*}[t]
\centering
\includegraphics[width=.95\linewidth]{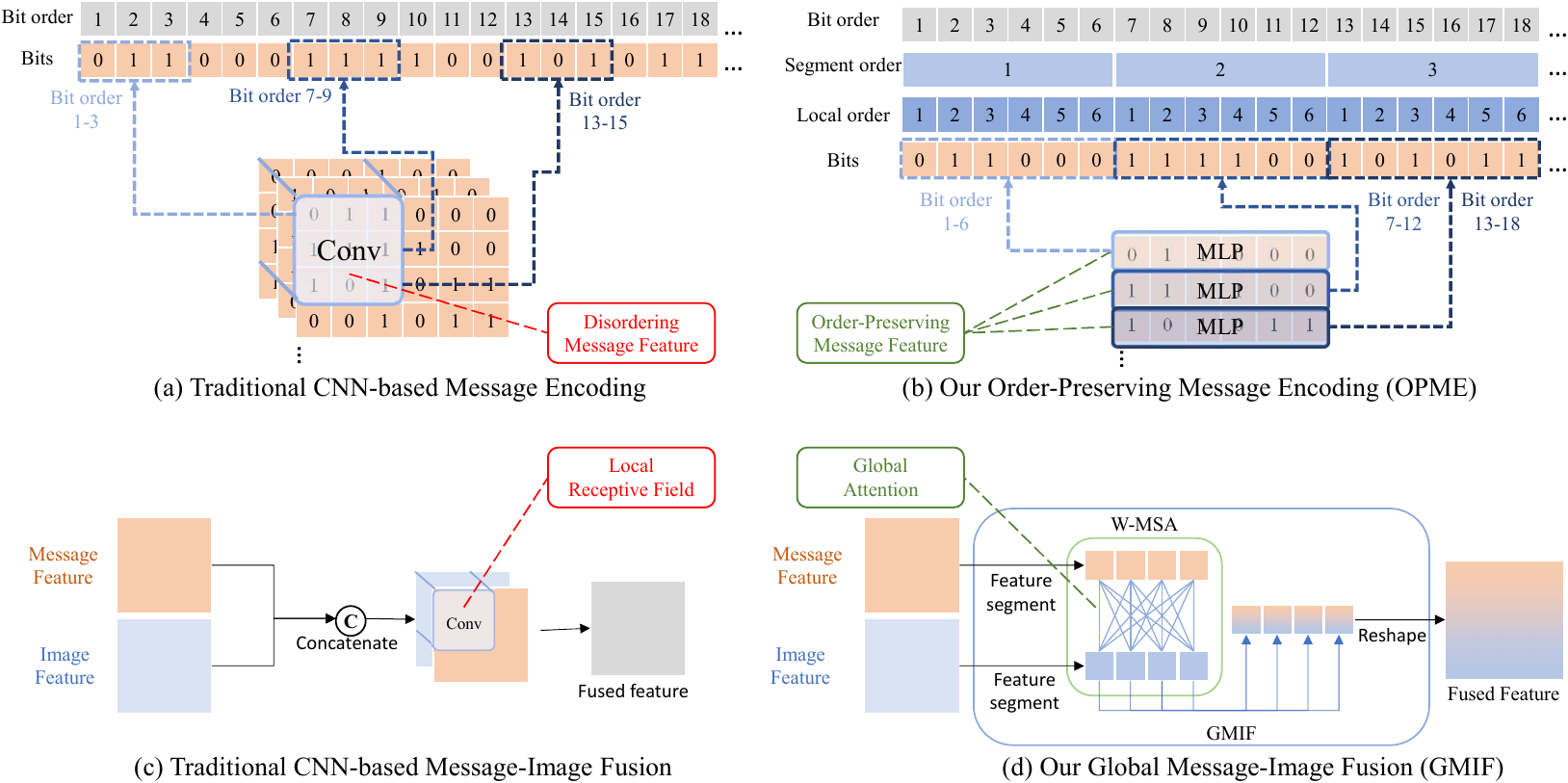} \\
% \vspace{-5pt}
\caption{The illustration of our proposed methods.\
(a) traditional CNN-based message encoding and (b) our proposed MLP-based message encoding:~\opmefull~(\opme) compares the difference in message encoding. (c) traditional feature fusion and (d) our proposed~\gmiffull~(\gmif) is the difference in message-image fusion.
}
\label{fig:compare_message_encoding}
\vspace{-5pt}
\end{figure*}

Traditional message hiding methods often rely on heuristic approaches, such as the Least Significant Bit method~\cite{mielikainen2006lsb}, which modifies individual pixels of a digital image to conceal message bits.
Recently, with the advancements of deep learning, latest methods have significantly improved the message capacity and imperceptibility~\cite{zhang2019steganogan,yu2020attention,tan2021channel,wei2022generative,yin2023anti}. These methods involve aligning and concatenating message bits with the cover image in a spatial manner while utilizing Convolutional Neural Network (CNN) to seamlessly integrate the message and image.

However, CNN-based approaches often yield suboptimal results in message encoding and recovery, particularly when employing convolutional operations to encode multiple layers of message bits as illustrated in \cref{fig:compare_message_encoding}(a). We have identified two main limitations in current CNN-based methods. Firstly, the indiscriminate use of convolutional kernels to encode message bits disrupts the sequential order of the original message bits. Secondly, the element-wise concatenation of message and image features, followed by convolutional operations, underestimates the substantial disparity between these two modalities, resulting in inadequate fusion as shown in \cref{fig:compare_message_encoding}(c).
% designed for spatially correlated data~\cite{lecun2015deep}, face limitations when applied to independently distributed secret message bits with a significant order. 

\begin{figure}[h]
\centering
\includegraphics[width=0.9\columnwidth]{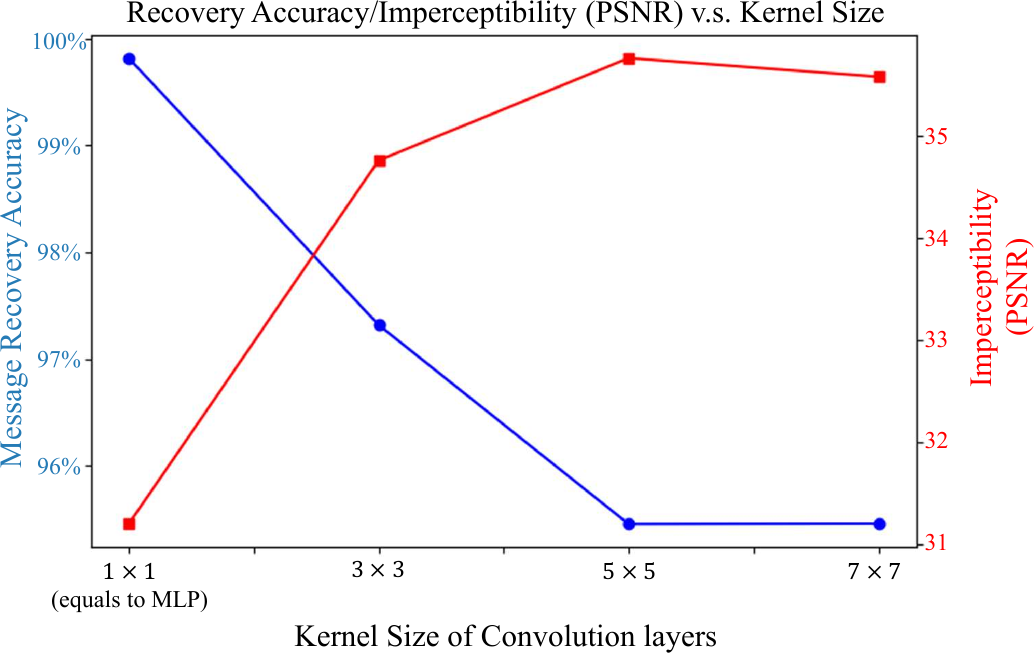}
% \vspace{-5pt}
\caption{The change in message recovery accuracy and stego image quality was investigated across different kernel sizes using the traditional CNN-based method SteganoGAN. The results reveal that the CNN-based structure struggles to enhance message recovery accuracy while maintaining imperceptibility.}
\label{fig:kernel_size_quality_acc}
\vspace{-10pt}
\end{figure}

To validate these assumptions, we gradually modified the kernel size of SteganoGAN~\cite{zhang2019steganogan}~ and designed experiments at a 4 BPP message capacity using Div2k~\cite{agustsson2017ntire}~dataset. As depicted in \cref{fig:kernel_size_quality_acc}, reducing the kernel size enhances the model's message recovery accuracy. With a decrease in kernel size from 7 to 1, 2D convolution is simplified to linear layer, focusing on 1D message segments along the channel dimension of the message tensor and encoding these segments sequentially. This process maintains the sequential order of the original bit string. On the other hand, the stego image quality improves as the kernel size increases due to a larger receptive field introduces more global interactions between the message and image features.

To address these challenges, as shown in \cref{fig:compare_message_encoding}(b), we leverage Multi-Layer Perceptrons (MLP) to design a more appropriate method for message encoding. The MLP-based message encoding phase can focus on small and complete message segments individually. Later, the Multi-Head Self-Attention (\msa)~\cite{vaswani2017attention} and positional embeddings (PEs) help to globally encode the message features while explicitly adding sequential information to the encoded message features. For message and image fusion, as shown in \cref{fig:compare_message_encoding}(d), we use \wmsafull~(\wmsa) to address the limitation of receptive fields in the CNN-based structure, enabling global interactions between message and image features.

Our approach, named StegaFormer, consists of three proposed modules for enhanced message encoding, decoding, and message-image fusion, respectively. We design the first two modules, \opmefull~(\opme) and \opmdfull~(\opmd), to effectively handle message segments while maintaining their order. The~\opmd~follows the same methodology and has a symmetric structure with~\opme~to decode the message. Long message bits are segmented into equal-length short messages. Later, MLPs encode and decode them separately~\cite{o2017introduction}. Additionally, \opme~and \opmd~take full advantage of sequential order by incorporating PEs into the message features and globally encodes and decodes the features using~\msa. The last one,~\gmiffull~(\gmif) employs a \wmsa~to fuse the information from the secret message and cover image to the stego image globally, greatly minimizing perceptible artifacts in the stego image.

Comprehensive experiments conducted on real-world datasets demonstrate the superior performance of our model in terms of recovery accuracy, message capacity, and imperceptibility. Importantly, our method exhibits a significant improvement in recovery accuracy. For message capacities ranging from 1 to 4 BPP, \nom~maintains a message recovery accuracy exceeding $99\%$. Even for high-capacity messages, \nom~achieves a message recovery accuracy of over $96\%$ at 6 BPP, surpassing the state-of-the-art method that achieves a 3 BPP capacity but with an even lower recovery accuracy. Our contributions are summarized as follows:
\begin{itemize}
    \item We are the first to point out that CNN disorders message features and is not appropriate for message hiding; consequently, we transfer to an MLP-based approach design.
    \item We introduce two novel modules \opme~\&~\opmd, which enhance the message encoding and decoding process by incorporating sequential order into the message.
    \item We design \gmif, which leverages the global interaction of \wmsa~to effectively fuse secret message and cover image features to stego image features.
    \item Experimental results on real-world datasets demonstrate the remarkable superiority of our framework over existing state-of-the-art methods in terms of accuracy, capacity, and imperceptibility.
\end{itemize}

\begin{figure*}[h]
\centering
\includegraphics[width=.9\linewidth]{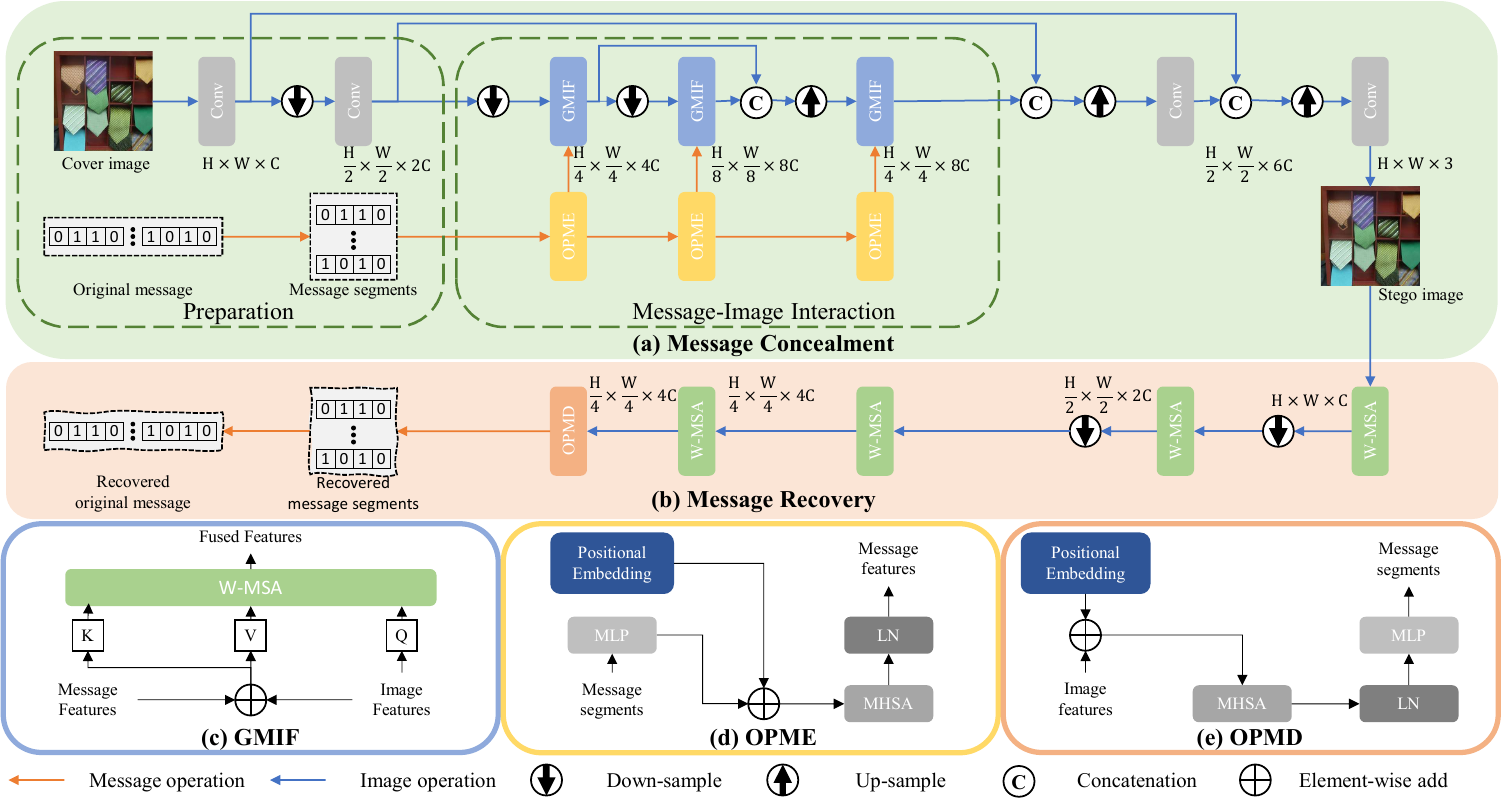}
% \vspace{-5pt}
% \caption{The overview of \nom~consisting of two main pipelines: (a) Message Concelment and (b) Message Recovery by three proposed components: (c)  \gmiffull~(\gmif), (d) \opmefull~(\opme) and (e) \opmdfull~(\opmd). The size indicated on top of the modules represents the shape of their respective outputs. We use LN to denote Layer Normalization and W-MSA to denote \wmsafull.}
\caption{The overview of \nom~ consists of two main pipelines: (a) Message Concealment and (b) Message Recovery, which are facilitated by three proposed components: (c) \gmiffull~(\gmif), (d) \opmefull~(\opme), and (e) \opmdfull~(\opmd). The sizes indicated above the modules represent the shapes of their respective outputs. LN denotes Layer Normalization, and W-MSA denotes \wmsafull.}
\label{fig:architecture}
\vspace{-10pt}
\end{figure*}

%%%%%%%%%%% related work
% reduce releated work
\section{Related Work}
\label{relwork}
\subsection{Message Hiding}
\label{sec:message_concealment}
Message hiding aims to embed confidential message bits within a cover image, resulting in a stego image, and recover message bits from the stego image.
Of course, it also needs to minimize any perceptible differences between cover image and stego image. Thus, the research history is around how to balance the three properties: message capacity, recovery accuracy, and imperceptibility.
 
Traditional methods often involve altering individual pixels in an image using heuristic techniques. They generally have a high recovery accuracy and imperceptibility, but low capacity. Specifically, early studies~\cite{mielikainen2006lsb,pevny2010using} manipulate the least significant bits of the cover image. Others~\cite{chandramouli2003image,bi2007robust,reddy2009high,holub2012designing} extend this approach to the frequency domain by transforming the cover image to different types of frequency domains for better performance. Despite their capacity for precise secret message recovery, these methods can only hide an extremely low BPP (\ie~0.4) to evade detection by steganalysis tools~\cite{goljan2006new,fridrich2012rich}.

Recently, CNN-based methods for message hiding greatly improve capacity and obtain significant attention~\cite{zhang2019steganogan,tan2021channel,chen2022learning,yin2023anti}. For instances, SteganoGAN~\cite{zhang2019steganogan} introduces 3D message tensor aligning in the spatial dimensions and ChatGAN~\cite{tan2021channel} incorporates channel attention to enhance accuracy and imperceptibility meanwhile keep a high capacity. SFTN~\cite{yin2023anti} is proposed to address the rounding error before generating the stego image. However, these CNN-based approaches contribute little to the improvement of message recovery accuracy.

In an effort to address the issue, other works~\cite{kishore2021fixed,chen2023learning} employ offline optimization, directly optimizing the stego image itself by iteratively adding adversarial perturbations to the stego image rather than improving the message hiding model. While this approach significantly enhances message recovery accuracy, optimizing individual stego images is time-consuming, and imperceptibility is greatly compromised.

Inspired by the difficulties encountered in previous works, we now turn our attention to the details of the message encoding procedure in these CNN-based models. We reveal the main constraint of CNN-based methods: CNNs disorder the message order, thus having limited accuracy. Thus, differing from previous CNN-based methods, we are the first to leverage MLPs with PEs to design an order-preserving framework.

\subsection{Multi-head Self-Attention}
\label{sec:vision_transformer}
\msa~mechanism is widely adopted in the domain of Natural Language Processing~\cite{ashish2017transformer} and various computer vision tasks~\cite{carion2020end,zhu2020deformable,ranftl2021vision,bai2022transfusion,li2022deepfusion}. The benefits of it stem from the dynamic weights and long-range encoding capability~\cite{khan2022transformers}. In our approach, we adopt the \wmsa~as the fundamental building block of the message-image fusion, enabling long contextual fusion of message and image features with less computational complexity.

% comparison w/ sota
\begin{table*}[htbp]
\centering
\scalebox{0.9}
{
\small
\begin{tabular}{l|ccc|ccc|ccc}
\toprule
Dataset    & \multicolumn{9}{c}{COCO} \\
\midrule
Metric      & \multicolumn{3}{c|}{ACC} & \multicolumn{3}{c|}{PSNR} & \multicolumn{3}{c}{SSIM} \\
\midrule
Capacity    & 1 BPP & 2 BPP & 3 BPP & 1 BPP & 2 BPP & 3 BPP & 1 BPP & 2 BPP & 3 BPP\\
\midrule
SteganoGAN~\cite{zhang2019steganogan} & 97.75$\%$ & 96.46$\%$ & 91.35$\%$ & 42.52 & 39.68 & 36.45 & 0.9894 & 0.9801 & 0.9650 \\
ChatGAN~\cite{tan2021channel} & 99.07\% & 97.46$\%$ & 96.18$\%$ & 46.42 & 43.17 & 41.84 & 0.9943 & 0.9880 & 0.9832 \\
\textbf{StegaFormer (Ours)} & \textbf{99.95$\%$} & \textbf{99.85$\%$} & \textbf{99.68$\%$} & \textbf{47.83} & \textbf{45.30} & \textbf{43.37} & \textbf{0.9969} & \textbf{0.9943} & \textbf{0.9914}  \\
% \midrule
% \rowcolor{gray!20}
% LISO~\cite{chen2023learning} & 100$\%$ & 100$\%$ & 100$\%$ & 33.42 & 31.80 & 26.98 & 0.9000 & 0.8700 & 0.7000 \\
\bottomrule
\toprule
Dataset    & \multicolumn{9}{c}{DIV2K} \\
\midrule
Metric     & \multicolumn{3}{c|}{ACC} & \multicolumn{3}{c|}{PSNR} & \multicolumn{3}{c}{SSIM} \\
\midrule
Capacity   & 1 BPP & 2 BPP & 3 BPP & 1 BPP & 2 BPP & 3 BPP & 1 BPP & 2 BPP & 3 BPP\\
\midrule
SteganoGAN~\cite{zhang2019steganogan} & 98.96$\%$ & 97.63$\%$ & 92.23$\%$ & 39.36 & 37.97 & 36.14 & 0.9793 & 0.8916 & 0.8343 \\
ChatGAN~\cite{tan2021channel} & 99.73$\%$ & 98.63$\%$ & 94.75$\%$ & 45.00 & 42.05 & 40.63 & 0.9929 & 0.9861 & 0.9782 \\
\textbf{StegaFormer (Ours)}  & \textbf{99.94$\%$} & \textbf{99.87$\%$} & \textbf{99.76$\%$}  & \textbf{51.98} & \textbf{49.07} & \textbf{47.31} & \textbf{0.9985} & \textbf{0.9970} & \textbf{0.9958} \\
% \midrule
% \rowcolor{gray!20}
% LISO~\cite{chen2023learning} & 100$\%$ & 100$\%$ & 100$\%$ & 33.12 & 32.77 & 29.58 & 0.8900 & 0.8900 & 0.8200 \\
\bottomrule
\end{tabular}
\vspace{-5pt}
}
\caption{Quantitative comparisons of SteganoGAN, ChatGAN, and our method on the COCO and DIV2K datasets are presented, with the best results highlighted in bold. ACC denotes message recovery accuracy. PSNR and SSIM are two metrics to measure the level of imperceptibility.}
% Additionally, we adding the SOTA offline method , LISO, to this compression.
\label{tab:bencharmerk_with_SOTA}
\vspace{-10pt}
\end{table*}

%%%%%%%%%%% methods
\section{Methods}
\label{sec:method}
As illustrated in \cref{fig:architecture}, our proposed method encompasses two pipeline for message concealment, i.e., \cref{fig:architecture}(a), and message recovery, i.e., \cref{fig:architecture}(b), respectively.
This approach sets itself apart from previous works by incorporating three crucial components: the~\gmif, as depicted in \cref{fig:architecture}(c), for the fusion between features from the secret message and cover image, the~\opme, as shown in \cref{fig:architecture}(d), for order-preserving message encoding,~\opmd~as shown in \cref{fig:architecture}(e) for corresponding message decoding.
% The \opmd~ is symmetric with \opme~to keep the same order-preserving design.

\subsection{Preparation}
\label{sec:prepare_message_image}
In the message concealment, cover image and secret message are pre-processed before the message-image interaction.

\subsubsection{Message Layout Preparation}
\label{sec:prepare_message}
% \noindent\textbf{Message Layout Preparation.}
A secret message $M \in \left\{0,1\right\}^{L}$, consisting of $L$ bits, is structured in the form $M \in \left\{0,1\right\}^{H \times W \times D}$ in previous approaches~\cite{zhang2019steganogan,tan2021channel,chen2022learning,yin2023anti}. Here, $H$ and $W$ represent the spatial dimensions of the message tensor, which correspond to the height and width of the cover image, while $D$ denotes the message capacity (e.g., $D=1$ for 1 BPP message capacity). Although this message layout can be easily processed using convolutional layers, it does introduce inappropriate limitations, as discussed in \cref{sec:intro}.

To address the issue, we introduce an unique message layout, where the message $M \in \left\{0,N_{r} \right\}^{L}$ is reshaped into $M_{\text{seg}} \in \left\{0,N_{r} \right\}^{N_{\text{ms}} \times L_{\text{ms}}}$.
Here, $N_{r}$ denotes the range of message elements (unless explicitly stated otherwise, $N_{r}$ is set to $1$) and $N_{\text{ms}}$ represents the number of message segments, $L_{\text{ms}}$ represents the length of each message segments, and $L_{\text{ms}} \ll L$. Considering the multi-level structure of \nom, we configure $N_{\text{ms}}$ and $L_{\text{ms}}$ at different sizes to align with the shape of the corresponding image features. 

% For example, in previous approaches, a cover image with dimensions $H=256$ and $W=256$ would accommodate a 1 BPP secret message as a message tensor with a shape of $256\times256$. However, in our method, we organize the message tensor as $4096\times16$ for the first \opme~to encode it. Here, $N_{\text{ms}}=4096$ and $L_{\text{ms}}=16$.

% \subsubsection{Preparation of Image Feature}
% \label{sec:prepare_image_feature}
% \noindent\textbf{Image Feature Preparation.}
\subsubsection{Image Feature Preparation}
\label{sec:prepare_image}
Given a cover image $I_{\text{cover}} \in \mathbb{R}^{H \times W \times 3}$, where $H$ and $W$ represent the height and width of the cover image, 
we employ two stages of operations to extract image feature $F_{\text{im}} \in \mathbb{R}^{(H/2^i \times W/2^i) \times 2^i \cdot C}$. Here, $i=1,2$ indicates the number of stages, and $C$ is the number of channels in the output feature. For models with different message capacities, we set $C=2\times L_{\text{ms}}$. Each stage consists of 2D convolution layers followed by a down-sampling operation. The down-sampling operation applies convolution to reduce the spatial size of the feature map by half while doubling its feature depth. The output of the image feature preparation phase is $F_{\text{im}} \in \mathbb{R}^{(H/4 \times W/4) \times 4C}$.
% after the two stages of feature extraction. 

\subsection{Order-Preserving Modules}
\subsubsection{\opmefull}
% \noindent\textbf{\opmefull.}
The main challenge in message encoding lies in effectively converting secret message bits into message features while retaining the sequential information of the message. To address this, we propose an \opme~that efficiently encodes the message while preserving both local and global sequential information. In this context, ``local" refers to the order within a message segment, while ``global" refers to the sequential order of the message segments. Our model consists of three \opme~in total to enable multi-scale interaction between the message and image features.

Each \opme~incorporates an MLP, referred to as $\text{MLP}_{\text{enc}}$, which functions as a channel encoder. Its purpose is to encode message segments $M_{\text{seg}}\in\left\{0,1\right\}^{N_{\text{ms}}\times L_{\text{ms}}}$ into message feature $F_{\text{msg}}$ following \cref{eq:channel_coding}:
% \begin{small}
\begin{align}
    F_{\text{msg}} &= \text{MLP}_{\text{enc}}(M_{\text{seg}}),
    \label{eq:channel_coding}
\end{align}
% \end{small}
where the shape of $F_{\text{msg}}$ is equal to the shape of $F_{\text{im}}$ at the corresponding \gmif. This learning-based channel encoder~\cite{o2017introduction} introduces redundancy into the original message segments, facilitating the stable transmission of the message feature within the model.

The importance of maintaining the sequential order of message segments is highlighted by the incorporation of PEs into the encoded segments. This is demonstrated in \cref{eq:global_message_encoding}:
% \begin{small}
\begin{align}
    F_{\text{msg}}^{\text{global}} &= \text{LN}(\text{MHSA}(F_{\text{msg}} + E_{\text{pos}})),
    \label{eq:global_message_encoding}
\end{align}
% \end{small}
where positional embedding $E_{\text{pos}}$ is introduced to the $F_{\text{msg}}$ followed by a~\msa~module~\cite{ashish2017transformer} to globally encode $F_{\text{msg}}$ to $F_{\text{msg}}^{\text{global}}$. Layer Normalization (LN)~\cite{ba2016layer} is employed to normalize the global message features.

\subsubsection{\opmdfull}
% \noindent\textbf{\opmdfull.}
\label{sec:message_decoder}
% \noindent\textbf{Message Decoder.}
As depicted in \cref{fig:architecture}(b), we employ a standard SwinTransformer~\cite{liu2021swint} with minor modifications to extract features from $I_{\text{stego}}$. Specifically, the down-sampling operations in the last two layers of the \wmsa~module within SwinTransformer are removed to match the spatial size to the feature of the secret message in the \opme. Subsequently, the \opmd~reconstructs the secret message segments $\hat{M}$ from the extracted features. The \opmd~follows the same methodology as the \opme, but in reverse order. 

\subsection{Global Message-Image Fusion}
Our \gmif, as show in \cref{eq:image_message_translation}:
% \begin{small}
\begin{align}
    F_{\text{stego}} &= \text{GMIF}(F_{\text{msg}}^{\text{global}}, F_{\text{im}}),
    \label{eq:image_message_translation}
\end{align}
% \end{small}
is designed to accomplish two fundamental goals of message hiding. These goals encompass: (1) the stego image should contain all the information from both the secret message and the cover image; (2) the ideal stego image should be identical to the cover image. The key, value and query in the attention mechanism are configured according to these two goals to translate the message and image information to the steganographic features $F_{\text{stego}}$:
% \begin{small}
\begin{align}
    K &= (F_{\text{im}} + F_{\text{msg}}^{\text{global}})W^{K}, \label{eq:k}\\
    V &= (F_{\text{im}} + F_{\text{msg}}^{\text{global}})W^{V}, \label{eq:v}\\
    Q &= F_{\text{im}}W^{Q}.
    \label{eq:q}
\end{align}
% \end{small}

The configuration of query, key and value utilized in our approach are inspired by the Transformer-based architecture commonly employed in language translation tasks. Drawing a parallel, we apply a similar mechanism to the task of message hiding sharing, as both involve dealing with a lack of spatial and semantic alignment between the secret message and cover image. The \gmif~effectively fuses the optimal stenographic information, which is the sum of $F_{\text{msg}}^{\text{global}}$ and $F_{\text{im}}$, into the desired feature representation of the stego image $F_{\text{stego}}$. Further insights into the various combinations of key, value, and query are provided in the supplementary material.

The basic building block of \gmif~is Windowed Multi-Head Self-Attention Layers (\wmsa)~\cite{liu2021swint}. One layer of \wmsa~with large window size (all window sizes equal to 16) is applied to ensure a comprehensive global interaction between message and image features in the \gmif.

As shown in \cref{eq:residual_reconstruction}, we apply up-sampling and convolution layers to the concatenated $F_{\text{stego}}$ and $F_{\text{im}}$:  
% \begin{small}
\begin{align}
    I_{\text{res}} = \text{Conv}(\text{Up-Sample}(\text{{Concat}}(F_{\text{stego}}, F_{\text{im}}))),
    \label{eq:residual_reconstruction}
\end{align}
% \end{small}
to reconstruct the residual image $I_\text{res}$. The up-sampling layer uses a de-convolution operation to double the height and width of the feature map and reduce the feature dimension by half. The final output of our model is $I_\text{stego}$, which is equal to the sum of $I_{\text{cover}}$ and $I_{\text{res}}$. Please check supplementary material for more detailed configurations and number of parameters of our model.

\subsection{Loss Function}
\label{sec:loss_function}
The objectives of message hiding are two-fold: to generate a stego image that closely resembles the cover image and to precisely retrieve the secret message from the stego image. In pursuit of these aims, we employ the loss function as defined in \cref{eq:total_loss}:
% \begin{small}
\begin{align}
    \mathcal{L}_{\text{total}} &= \mathcal{L}_{\text{img}} + \mathcal{L}_{\text{msg}}, \label{eq:total_loss}\\
    \mathcal{L}_{\text{img}} &= \lambda_{1}\mathcal{L}_{\text{MSE}}(I_{\text{cover}},I_{\text{stego}}) + \lambda_{2}\mathcal{L}_{\text{LPIPs}}(I_{\text{cover}},I_{\text{stego}}),\label{eq:image_loss}\\
    \mathcal{L}_{\text{msg}} &=
    \begin{cases}
        \mathcal{L}_{\text{BCE}}(M, \hat{M}), & \text{if} ~N_{r}=1\\
        \mathcal{L}_{\text{MSE}}(M, \hat{M}). & \text{if} ~N_{r}>1
    \end{cases} \label{eq:msg_loss}
\end{align}
% \end{small}

$\mathcal{L}_{\text{img}}$ in \cref{eq:image_loss} combines a weighted sum of mean square error loss and perceptual loss~\cite{zhang2018unreasonable}. We employ binary cross-entropy loss for $\mathcal{L}_{\text{msg}}$ in \cref{eq:msg_loss} when the range of message $N_{r}=1$ and $\mathcal{L}_{\text{msg}}$ change to mean square error loss when the message range $N_{r}>1$.

%%%%%%%%%%% experiments
\section{Experiments}
\label{sec:exp}
To demonstrate the effectiveness of our proposed approach, we conduct a comparative analysis of message recovery accuracy and imperceptibility against two state-of-the-art message hiding methods: SteganoGAN~\cite{zhang2019steganogan} and ChatGAN~\cite{tan2021channel}. This analysis is carried out across different message capacity levels. 

% benchmarks
% \subsection{Experimental Settings}
% \label{sec:exp_setting}
\noindent\textbf{Experimental Settings.}
Following previous studies~\cite{zhang2019steganogan,tan2021channel}, our method is evaluated on two public datasets: COCO~\cite{lin2014microsoft} and DIV2K~\cite{agustsson2017ntire}. We chose the COCO dataset due to its inclusion of diverse scenes and rich contextual image information, while the DIV2K dataset consists of high-resolution images. The COCO dataset for message hiding contains 25,000 images randomly sampled from the original COCO training split for model training, as well as 500 images randomly sampled from the original COCO test split for model evaluation. The DIV2K dataset contains 800 training images and 100 evaluation images. Throughout both the training and evaluation phases, we fix the image size to $256\times256$ pixels for both datasets. The training process for each model consists of 100,000 iterations, with a batch size of 2. We set $\lambda_{1}=1 \times 10^{-4}$ and $\lambda_{2}=1 \times 10^{-6}$ to balance these different losses. More detailed configurations of our model are listed in the supplementary material due to the page limitation.

% \subsection{Evaluation Metrics}
% \label{sec:eval_metrics}
\noindent\textbf{Evaluation Metrics.}
Three metrics are used to evaluate the model performance. The message recovery accuracy is utilized to assess the percentage of successfully recovered secret messages. The imperceptibility of stego images is measured using the PSNR and SSIM~\cite{wang2004image}.

\subsection{Comparison with State-of-the-Art Methods}
\label{sec:compare_sota}
A comprehensive comparison between our method and two existing state-of-the-art methods~\cite{zhang2019steganogan,tan2021channel} is presented in~\cref{tab:bencharmerk_with_SOTA}. Our method consistently outperforms the other methods in terms of recovery accuracy and imperceptibility at 1,2,3 BPP message capacity.

Specifically, at a 3 BPP capacity, our method achieves an impressive recovery accuracy of $99.68\%$ and $99.76\%$ for the COCO and DIV2K datasets, respectively. This represents a significant improvement of $3.5\%$ and $5.0\%$ compared to the current state-of-the-art accuracies achieved by ChatGAN~\cite{tan2021channel}. These improvements demonstrate the effectiveness of~\opme~in achieving high-accuracy message hiding.

Furthermore, our method enhances imperceptibility in terms of both PSNR and SSIM across different message capacities. For example, compared with ChatGAN~\cite{tan2021channel} at a 3 BPP message capacity, our method improves the PSNR from 41.84 to 43.37 on the COCO dataset, and from 40.63 to 47.31 on the DIV2K dataset. These results underscore the effectiveness of~\gmif~in fusion of steganographic information into stego images.

% Additionally, we also include a comparison between our method and LISO~\cite{chen2023learning}. As mentioned in Section~\ref{sec:message_concealment}, this approach does not prioritize message hiding models but rather focuses on offline optimization of individual stego images. While it significantly enhances recovery accuracy, it compromises imperceptibility and does not enhance the performance of the message hiding model.

\cref{fig:msg_hiding_img_quality_and_residual}(b) displays residual images generated by our method and previous methods~\cite{zhang2019steganogan,tan2021channel} at 3 BPP message capacity. Residual images of our method demonstrate that minimal changes have been added to  cover images. In contrast, previous approaches spread higher residuals all over the cover image, leading to lower imperceptibility. More qualitative results are in the supplementary material.

\subsection{Models for High Message Capacity}
\label{sec:large_message_capacity_model}

\begin{table}[h]
\vspace{-5pt}
\centering
\scalebox{0.9}
{
\begin{tabular}{l|cccccc}
\toprule
BPP & 4 & 6 & 8\\
\midrule
ACC &  99.27$\%$ & 96.65$\%$ & 91.78$\%$\\
PSNR & 41.87 & 40.37 & 34.70 \\
SSIM & 0.9877 & 0.9803 &  0.9508\\
\bottomrule
\end{tabular}
}
\caption{Quantitative results of~\nom~with message capacities at 4, 6 and 8 BPP on the COCO dataset.}
\label{tab:maximum_message_capacity}
\vspace{-5pt}
\end{table}

\begin{figure*}[htbp]
	\begin{minipage}[t]{0.5\linewidth}
		\centering
		\includegraphics[width=\linewidth]{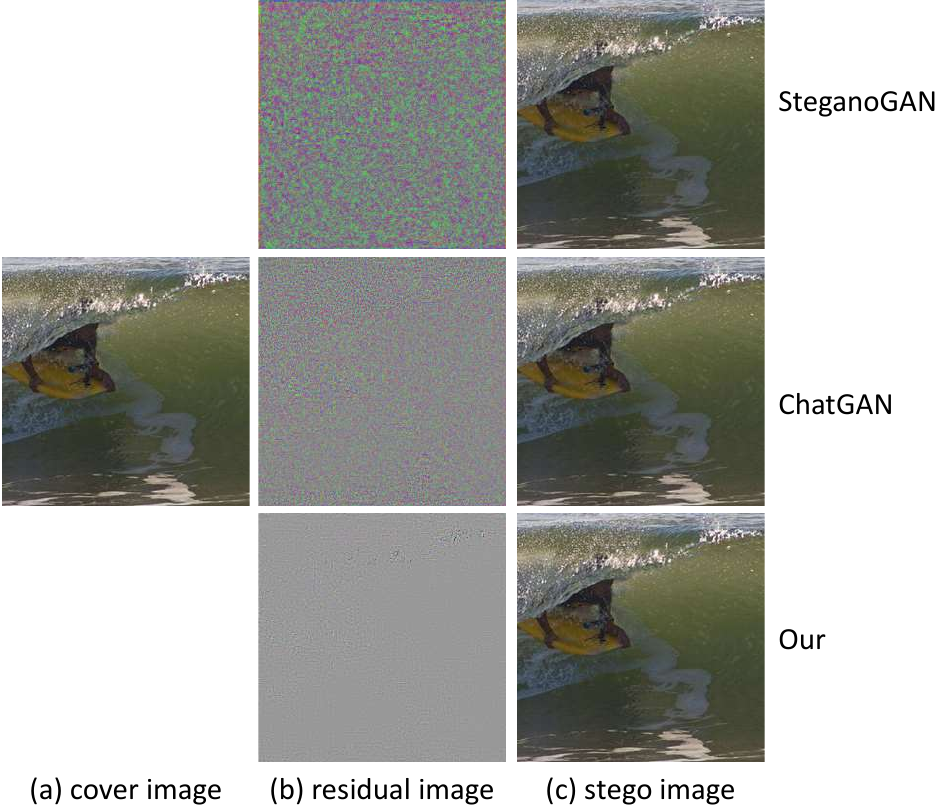}
		\caption{Qualitative results of 3 bits per pixel (BPP) message hiding using different methods: (a) cover image, (b) residual images, and (c) stego images. The images in the top row are obtained from SteganoGAN, those in the middle row from ChatGAN, and the bottom row showcases our method. For enhanced visualization, the residual images are multiplied by 100. Best viewed in digital version.}
        \label{fig:msg_hiding_img_quality_and_residual}
	\end{minipage}
 \quad
	\begin{minipage}[t]{0.5\linewidth}
		\centering
		\includegraphics[width=0.9\linewidth]{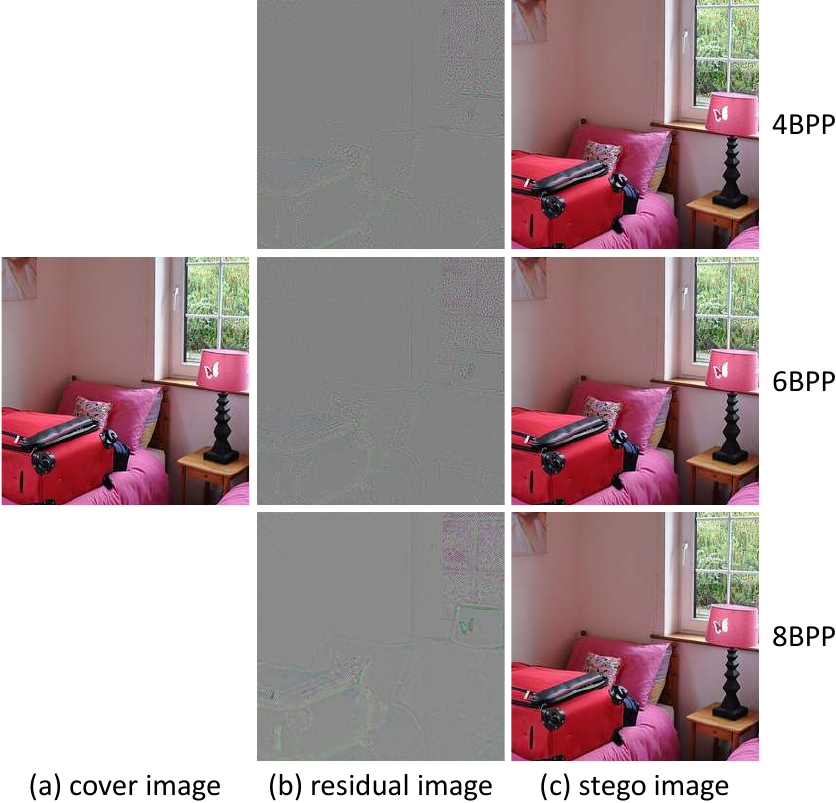}
		\caption{Qualitative results of high-capacity message hiding with our method: (a) cover image, (b) residual images, and (c) stego images. Samples at 4 BPP, 6 BPP, and 8 BPP message capacities are displayed in the first, second, and third rows, respectively. The residual images are multiplied by 100 for visualization. Best viewed in digital version.}
        \label{fig:large_capacity_img_quality_and_residual}
	\end{minipage}
\vspace{-20pt}
\end{figure*}

We conduct a series of experiments to showcase the performance of our method in handling high message capacities. As shown in~\cref{tab:maximum_message_capacity}, the results demonstrate that our method achieves a message recovery accuracy exceeding $90\%$ for message capacities up to 8 BPP. Compared with existing methods, these results highlight the remarkable improvement achieved by our method in both message capacity and recovery accuracy.
Besides, \cref{fig:large_capacity_img_quality_and_residual}(b)(c) depicts the residual  and stego images by our models with high message capacities. The residual images of these high-capacity models demonstrate the outstanding imperceptibility of our models.

There are two approaches to increase the message hiding capacity in our method: increasing the length of message segments $L_{\text{ms}}$ and expanding the range of message elements $N_{r}$. The 4 BPP model has $(L_{\text{ms}}, N_{r}) = (64, 1)$, the 6 BPP model has $(L_{\text{s}}, N_{r}) = (48, 3)$, and the 8 BPP model has $(L_{\text{ms}}, N_{r}) = (32, 15)$. More details of $N_{r}$ and $L_{\text{ms}}$ are provided in the supplementary material.

% details of N_{r} and L_{s} move to supplementary material
% When $N_{r}=1$, typical choices for $L_{\text{s}}$ include 16, 32, 48, and 64, corresponding to message capacities of 1, 2, 3, and 4 BPP, respectively. $N_{r}$ can also be increased to achieve higher message capacities. For example, $N_{r} = 2^i - 1, \quad \text{where } i = 2, 3, 4$, represents 2, 3, and 4 bits per message element, respectively. Combining these two factors can lead to a higher message capacity for our models. The listed capacities in \cref{tab:maximum_message_capacity} are achieved through combinations of symbol ranges $N_{r}$ and message segment lengths $L_{\text{s}}$. 

\subsection{Steganalysis}
\label{sec:stego_analysis}
Steganalysis methods~\cite{boroumand2018deep,zhang2019depth,you2020siamese} play a crucial role in assessing the security of stego images. Typically, this involves utilizing a machine learning model to classify images as either cover images or stego images. In our study, we choose the state-of-the-art steganalysis approach SiaStegNet~\cite{you2020siamese} to conduct a comprehensive benchmark comparison between our method and existing methods~\cite{zhang2019steganogan,tan2021channel}. The SiaStegNet models are trained and evaluated on the COCO dataset.

\begin{figure}[h]
\centering
\includegraphics[width=0.8\columnwidth]{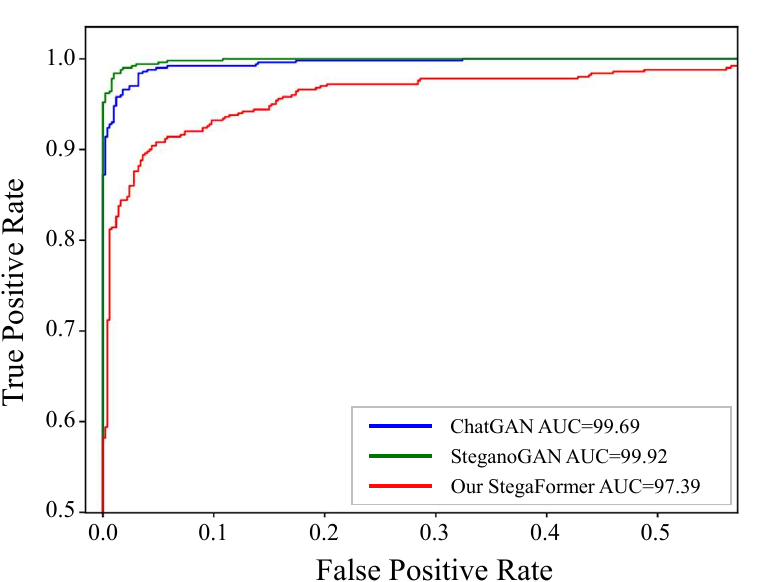}
\caption{The comparison of the detection accuracy of SiaStegNet models for ChatGAN, SteganoGAN and our approach. Lower AUC represents the stego images by our~\nom~are more difficult to detect than the other two methods. Our approach shows lowest AUC.}
\label{fig:steganalysis}
\vspace{-15pt}
\end{figure}

\cref{fig:steganalysis} presents the steganalysis results, demonstrating that our model achieves the lowest Area Under the Curve (AUC) among the three compared methods. These results indicates that our method exhibits the lowest detection rate and has the best security of stego images. All message hiding models for comparison have a 3 BPP message capacity.

\begin{figure}[h]
\centering
\includegraphics[width=0.85\columnwidth]{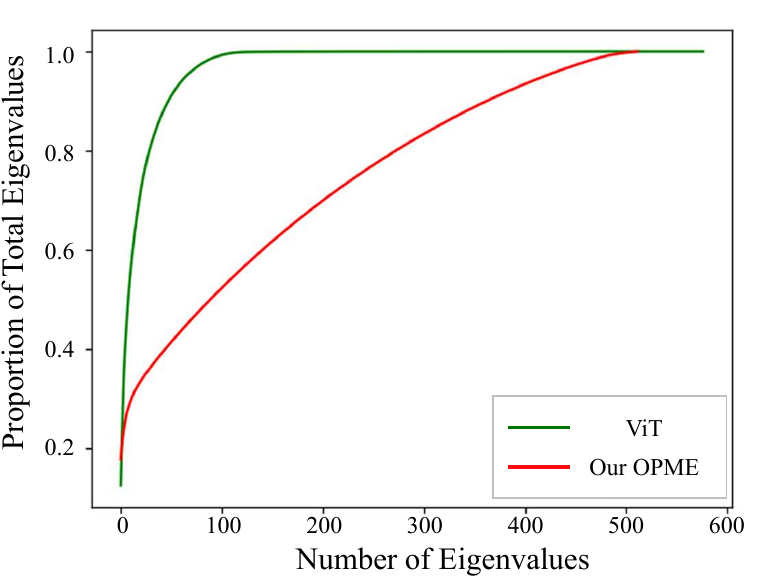}
\caption{The comparison of the ratio between the cumulative sum of the top n eigenvalues and the sum of all eigenvalues. Green for the PEs from pre-trained ViT and red for the PEs from~\opme. The PEs from \opme~contain more independent positional information.}
\label{fig:pose_compare_accumulated_eigen}
\vspace{-15pt}
\end{figure}

\subsection{Ablation Study}
\label{sec:ablation}
To prove the effectiveness of our purposed components. We first evaluate the performance of \opme, \opmd~and \gmif. Later, go into the details of the \opme~and \opmd.
\subsubsection{Effectiveness of~\opme~,~\opmd~and~\gmif}
\label{sec:ablation_global_message_encoder}

As illustrated in ~\cref{tab:ablate_OPME_OPMD_GMIF}, to assess the effectiveness of our order-preserving in~\opme~and~\opmd~and global interaction design in~\gmif, we scale down our model by removing the~\opme~and~\opmd~and~\gmif~step by step to set up a baseline model, then perform the message hiding training at a 4 BPP message payload. In their absence, these components are replaced with 2D convolution layers. It is evident that the absence of both~\opme~and~\opmd~and~\gmif~prevents the model from successfully hiding the secret message within the cover image. A model using only~\gmif~can boost accuracy via more effective message-image fusion. The involvement of~\opme~and~\opmd~significantly increases the accuracy of message recovery, highlighting the effectiveness of preserving message sequential order in message encoding.

\begin{table}[htbp]
\centering
\scalebox{0.9}
{
\begin{tabular}{l|ccc}
\toprule
Models & ACC & PSNR & SSIM \\
\midrule
 baseline & 50.01\% & 53.4 & 0.9988\\
 +\gmif  & 87.60\% & 43.78 & 0.9908\\
 +\gmif+\opme+\opmd & 99.24\% & 41.78 & 0.9875\\
\bottomrule
\end{tabular}
}
\caption{Ablation of the proposed~\opme, \opmd~and~\gmif. We replace our proposed components with 2D convolution layers in the baseline model which disable the global message-image fusion and the order-preserving message encoding and decoding, respectively.}
\label{tab:ablate_OPME_OPMD_GMIF}
\vspace{-10pt}
\end{table}

\subsubsection{Effectiveness of PE within \opme~and \opmd}
\label{sec:ablation_global_message_encoder}

\begin{table}[h]
\vspace{-5pt}
\centering
\scalebox{0.9}
{
\begin{tabular}{lll|ccc}
\toprule
MLP & \msa & PE & ACC & PSNR & SSIM \\
\midrule
$\checkmark$ & $\times$ & $\times$ & 98.59\% & 40.59 & 0.9826\\
$\checkmark$ & $\checkmark$ & $\times$ & 98.68\% & 40.70 & 0.9833\\
$\checkmark$ & $\times$ & $\checkmark$ & 97.92\% &  40.05& 0.9786\\
$\checkmark$ & $\checkmark$ & $\checkmark$ & 99.24\% & 41.78 & 0.9875\\
\bottomrule
\end{tabular}
}
\caption{Ablation of the proposed~\opme~and \opmd. We compare the performance of different version of them, which gradually add the PE and \msa~to validate the effectiveness of these components.}
\label{tab:ablate_opme}
\vspace{-5pt}
\end{table}

As illustrated in ~\cref{tab:ablate_opme}, we assess the efficacy of the features of~\opme~and \opmd~at 4 BPP using COCO dataset. We start with the model only contains MLP as a message encoder and decoder. The addition of solely positional embedding or \msa~after the MLP module results in sub-optimal performance. The best performance is achieved with the default design of our~\opme~and \opmd, which include MLP, \msa, and PEs.

To further validate whether~\opme~and \opmd~appropriately models the independent nature of message segments, we assess the positional embeddings learned by~\opme. More specifically, we compare the characteristics of positional embeddings obtained from ViT with those from~\opme. As illustrated in ~\cref{fig:pose_compare_accumulated_eigen}, the positional embeddings obtained from ViT exhibit a prevalence of significant eigenvalues, suggesting that numerous image patches share similar positional information for image classification task. In contrast, the positional embeddings acquired from~\opme~do not exhibit prominent eigenvalues, indicating that the positional embeddings in~\opme~contain more independent positional information than those in ViT. This demonstrates that~\opme~has effectively learned to assign distinct positional information to each individual message segment.

\section{Conclusion}
\label{conclusion}
In this paper, we propose~\nom, a novel message hiding framework for high accuracy message concealment and recovery.~\nom~is able to significantly increase the message recovery accuracy by emphasizing the order of message bits and cross-modality feature fusion. Experiments on real-world datasets demonstrate that our method consistently outperforms the state-of-the-art methods in terms of message recovery accuracy, message capacity, and imperceptibility.

\appendix

\section*{Ethical Statement}

There are no ethical issues.

\section*{Acknowledgments}
We thank Wang Wenfeng for conducting early experiments and Lu Lvjian for reviewing the manuscript.

\bibliographystyle{ijcai24}
\bibliography{EMH}

\begin{thebibliography}{}

\bibitem[\protect\citeauthoryear{Agustsson and Timofte}{2017}]{agustsson2017ntire}
Eirikur Agustsson and Radu Timofte.
\newblock Ntire 2017 challenge on single image super-resolution: Dataset and study.
\newblock In {\em Proceedings of the IEEE conference on computer vision and pattern recognition workshops}, pages 126--135, 2017.

\bibitem[\protect\citeauthoryear{Ba \bgroup \em et al.\egroup }{2016}]{ba2016layer}
Jimmy~Lei Ba, Jamie~Ryan Kiros, and Geoffrey~E Hinton.
\newblock Layer normalization.
\newblock {\em arXiv preprint arXiv:1607.06450}, 2016.

\bibitem[\protect\citeauthoryear{Bai \bgroup \em et al.\egroup }{2022}]{bai2022transfusion}
Xuyang Bai, Zeyu Hu, Xinge Zhu, Qingqiu Huang, Yilun Chen, Hongbo Fu, and Chiew-Lan Tai.
\newblock Transfusion: Robust lidar-camera fusion for 3d object detection with transformers.
\newblock In {\em Proceedings of the IEEE/CVF conference on computer vision and pattern recognition}, pages 1090--1099, 2022.

\bibitem[\protect\citeauthoryear{Bi \bgroup \em et al.\egroup }{2007}]{bi2007robust}
Ning Bi, Qiyu Sun, Daren Huang, Zhihua Yang, and Jiwu Huang.
\newblock Robust image watermarking based on multiband wavelets and empirical mode decomposition.
\newblock {\em IEEE Transactions on Image Processing}, 16(8):1956--1966, 2007.

\bibitem[\protect\citeauthoryear{Boroumand \bgroup \em et al.\egroup }{2018}]{boroumand2018deep}
Mehdi Boroumand, Mo~Chen, and Jessica Fridrich.
\newblock Deep residual network for steganalysis of digital images.
\newblock {\em IEEE Transactions on Information Forensics and Security}, 14(5):1181--1193, 2018.

\bibitem[\protect\citeauthoryear{Carion \bgroup \em et al.\egroup }{2020}]{carion2020end}
Nicolas Carion, Francisco Massa, Gabriel Synnaeve, Nicolas Usunier, Alexander Kirillov, and Sergey Zagoruyko.
\newblock End-to-end object detection with transformers.
\newblock In {\em ECCV}, pages 213--229. Springer, 2020.

\bibitem[\protect\citeauthoryear{Chandramouli \bgroup \em et al.\egroup }{2003}]{chandramouli2003image}
Rajarathnam Chandramouli, Mehdi Kharrazi, and Nasir Memon.
\newblock Image steganography and steganalysis: Concepts and practice.
\newblock In {\em International Workshop on Digital Watermarking}, pages 35--49. Springer, 2003.

\bibitem[\protect\citeauthoryear{Chen \bgroup \em et al.\egroup }{2022}]{chen2022learning}
Xiangyu Chen, Varsha Kishore, and Kilian~Q Weinberger.
\newblock Learning iterative neural optimizers for image steganography.
\newblock In {\em The Eleventh International Conference on Learning Representations}, 2022.

\bibitem[\protect\citeauthoryear{Chen \bgroup \em et al.\egroup }{2023}]{chen2023learning}
Xiangyu Chen, Varsha Kishore, and Kilian~Q Weinberger.
\newblock Learning iterative neural optimizers for image steganography.
\newblock {\em arXiv preprint arXiv:2303.16206}, 2023.

\bibitem[\protect\citeauthoryear{Fridrich and Kodovsky}{2012}]{fridrich2012rich}
Jessica Fridrich and Jan Kodovsky.
\newblock Rich models for steganalysis of digital images.
\newblock {\em IEEE Transactions on information Forensics and Security}, 7(3):868--882, 2012.

\bibitem[\protect\citeauthoryear{Goljan \bgroup \em et al.\egroup }{2006}]{goljan2006new}
Miroslav Goljan, Jessica Fridrich, and Taras Holotyak.
\newblock New blind steganalysis and its implications.
\newblock In {\em Security, Steganography, and Watermarking of Multimedia Contents VIII}, volume 6072, page 607201. International Society for Optics and Photonics, 2006.

\bibitem[\protect\citeauthoryear{Holub and Fridrich}{2012}]{holub2012designing}
Vojt{\v{e}}ch Holub and Jessica Fridrich.
\newblock Designing steganographic distortion using directional filters.
\newblock In {\em 2012 IEEE International workshop on information forensics and security (WIFS)}, pages 234--239. IEEE, 2012.

\bibitem[\protect\citeauthoryear{Khan \bgroup \em et al.\egroup }{2022}]{khan2022transformers}
Salman Khan, Muzammal Naseer, Munawar Hayat, Syed~Waqas Zamir, Fahad~Shahbaz Khan, and Mubarak Shah.
\newblock Transformers in vision: A survey.
\newblock {\em ACM computing surveys (CSUR)}, 54(10s):1--41, 2022.

\bibitem[\protect\citeauthoryear{Kishore \bgroup \em et al.\egroup }{2021}]{kishore2021fixed}
Varsha Kishore, Xiangyu Chen, Yan Wang, Boyi Li, and Kilian~Q Weinberger.
\newblock Fixed neural network steganography: Train the images, not the network.
\newblock In {\em International Conference on Learning Representations}, 2021.

\bibitem[\protect\citeauthoryear{Li \bgroup \em et al.\egroup }{2022}]{li2022deepfusion}
Yingwei Li, Adams~Wei Yu, Tianjian Meng, Ben Caine, Jiquan Ngiam, Daiyi Peng, Junyang Shen, Yifeng Lu, Denny Zhou, Quoc~V Le, et~al.
\newblock Deepfusion: Lidar-camera deep fusion for multi-modal 3d object detection.
\newblock In {\em Proceedings of the IEEE/CVF Conference on Computer Vision and Pattern Recognition}, pages 17182--17191, 2022.

\bibitem[\protect\citeauthoryear{Lin \bgroup \em et al.\egroup }{2014}]{lin2014microsoft}
Tsung-Yi Lin, Michael Maire, Serge Belongie, James Hays, Pietro Perona, Deva Ramanan, Piotr Doll{\'a}r, and C~Lawrence Zitnick.
\newblock Microsoft coco: Common objects in context.
\newblock In {\em European conference on computer vision}, pages 740--755. Springer, 2014.

\bibitem[\protect\citeauthoryear{Liu \bgroup \em et al.\egroup }{2021}]{liu2021swint}
Ze~Liu, Yutong Lin, Yue Cao, Han Hu, Yixuan Wei, Zheng Zhang, Stephen Lin, and Baining Guo.
\newblock Swin transformer: Hierarchical vision transformer using shifted windows.
\newblock In {\em ICCV}, pages 10012--10022, 2021.

\bibitem[\protect\citeauthoryear{Mielikainen}{2006}]{mielikainen2006lsb}
Jarno Mielikainen.
\newblock Lsb matching revisited.
\newblock {\em IEEE signal processing letters}, 13(5):285--287, 2006.

\bibitem[\protect\citeauthoryear{O’shea and Hoydis}{2017}]{o2017introduction}
Timothy O’shea and Jakob Hoydis.
\newblock An introduction to deep learning for the physical layer.
\newblock {\em IEEE Transactions on Cognitive Communications and Networking}, 3(4):563--575, 2017.

\bibitem[\protect\citeauthoryear{Pevn{\`y} \bgroup \em et al.\egroup }{2010}]{pevny2010using}
Tom{\'a}{\v{s}} Pevn{\`y}, Tom{\'a}{\v{s}} Filler, and Patrick Bas.
\newblock Using high-dimensional image models to perform highly undetectable steganography.
\newblock In {\em International Workshop on Information Hiding}, pages 161--177, 2010.

\bibitem[\protect\citeauthoryear{Ranftl \bgroup \em et al.\egroup }{2021}]{ranftl2021vision}
Ren{\'e} Ranftl, Alexey Bochkovskiy, and Vladlen Koltun.
\newblock Vision transformers for dense prediction.
\newblock In {\em ICCV}, pages 12179--12188, 2021.

\bibitem[\protect\citeauthoryear{Reddy and Raja}{2009}]{reddy2009high}
HS~Manjunatha Reddy and KB~Raja.
\newblock High capacity and security steganography using discrete wavelet transform.
\newblock {\em International Journal of Computer Science and Security (IJCSS)}, 3(6):462, 2009.

\bibitem[\protect\citeauthoryear{Tan \bgroup \em et al.\egroup }{2021}]{tan2021channel}
Jingxuan Tan, Xin Liao, Jiate Liu, Yun Cao, and Hongbo Jiang.
\newblock Channel attention image steganography with generative adversarial networks.
\newblock {\em IEEE Transactions on Network Science and Engineering}, 2021.

\bibitem[\protect\citeauthoryear{Tancik \bgroup \em et al.\egroup }{2020}]{2019stegastamp}
Matthew Tancik, Ben Mildenhall, and Ren Ng.
\newblock Stegastamp: Invisible hyperlinks in physical photographs.
\newblock In {\em CVPR}, 2020.

\bibitem[\protect\citeauthoryear{Vaswani \bgroup \em et al.\egroup }{2017a}]{ashish2017transformer}
Ashish Vaswani, Noam Shazeer, Niki Parmar, Jakob Uszkoreit, Llion Jones, Aidan~N Gomez, \L~ukasz Kaiser, and Illia Polosukhin.
\newblock Attention is all you need.
\newblock In {\em Advances in Neural Information Processing Systems}, volume~30, 2017.

\bibitem[\protect\citeauthoryear{Vaswani \bgroup \em et al.\egroup }{2017b}]{vaswani2017attention}
Ashish Vaswani, Noam Shazeer, Niki Parmar, Jakob Uszkoreit, Llion Jones, Aidan~N Gomez, {\L}ukasz Kaiser, and Illia Polosukhin.
\newblock Attention is all you need.
\newblock {\em Advances in neural information processing systems}, 30, 2017.

\bibitem[\protect\citeauthoryear{Wang \bgroup \em et al.\egroup }{2004}]{wang2004image}
Zhou Wang, Alan~C Bovik, Hamid~R Sheikh, and Eero~P Simoncelli.
\newblock Image quality assessment: from error visibility to structural similarity.
\newblock {\em IEEE transactions on image processing}, 13(4):600--612, 2004.

\bibitem[\protect\citeauthoryear{Wei \bgroup \em et al.\egroup }{2022}]{wei2022generative}
Ping Wei, Sheng Li, Xinpeng Zhang, Ge~Luo, Zhenxing Qian, and Qing Zhou.
\newblock Generative steganography network.
\newblock In {\em Proceedings of the 30th ACM International Conference on Multimedia}, pages 1621--1629, 2022.

\bibitem[\protect\citeauthoryear{Wengrowski and Dana}{2019}]{Wengrowski2019lfm}
Eric Wengrowski and Kristin Dana.
\newblock Light field messaging with deep photographic steganography.
\newblock In {\em CVPR}, 2019.

\bibitem[\protect\citeauthoryear{Yin \bgroup \em et al.\egroup }{2023}]{yin2023anti}
Xiaolin Yin, Shaowu Wu, Ke~Wang, Wei Lu, Yicong Zhou, and Jiwu Huang.
\newblock Anti-rounding image steganography with separable fine-tuned network.
\newblock {\em IEEE Transactions on Circuits and Systems for Video Technology}, 2023.

\bibitem[\protect\citeauthoryear{You \bgroup \em et al.\egroup }{2020}]{you2020siamese}
Weike You, Hong Zhang, and Xianfeng Zhao.
\newblock A siamese cnn for image steganalysis.
\newblock {\em IEEE Transactions on Information Forensics and Security}, 16:291--306, 2020.

\bibitem[\protect\citeauthoryear{Yu}{2020}]{yu2020attention}
Chong Yu.
\newblock Attention based data hiding with generative adversarial networks.
\newblock In {\em Proceedings of the AAAI conference on artificial intelligence}, volume~34, pages 1120--1128, 2020.

\bibitem[\protect\citeauthoryear{Zhang \bgroup \em et al.\egroup }{2018}]{zhang2018unreasonable}
Richard Zhang, Phillip Isola, Alexei~A Efros, Eli Shechtman, and Oliver Wang.
\newblock The unreasonable effectiveness of deep features as a perceptual metric.
\newblock In {\em CVPR}, pages 586--595, 2018.

\bibitem[\protect\citeauthoryear{Zhang \bgroup \em et al.\egroup }{2019a}]{zhang2019steganogan}
Kevin~Alex Zhang, Alfredo Cuesta-Infante, and Kalyan Veeramachaneni.
\newblock Steganogan: High capacity image steganography with gans.
\newblock {\em arXiv preprint arXiv:1901.03892}, 2019.

\bibitem[\protect\citeauthoryear{Zhang \bgroup \em et al.\egroup }{2019b}]{zhang2019depth}
Ru~Zhang, Feng Zhu, Jianyi Liu, and Gongshen Liu.
\newblock Depth-wise separable convolutions and multi-level pooling for an efficient spatial cnn-based steganalysis.
\newblock {\em IEEE Transactions on Information Forensics and Security}, 15:1138--1150, 2019.

\bibitem[\protect\citeauthoryear{Zhu \bgroup \em et al.\egroup }{2020}]{zhu2020deformable}
Xizhou Zhu, Weijie Su, Lewei Lu, Bin Li, Xiaogang Wang, and Jifeng Dai.
\newblock Deformable detr: Deformable transformers for end-to-end object detection.
\newblock In {\em International Conference on Learning Representations}, 2020.

\end{thebibliography}

\end{document}